\documentclass[runningheads]{llncs}
\usepackage{times}
\usepackage{latexsym}
\usepackage{url}
\usepackage{graphicx}
\usepackage{amsmath}
\usepackage{amssymb}
\usepackage{booktabs}
\usepackage{multirow}
\usepackage{diagbox}
\usepackage{adjustbox}
\usepackage{xspace}
\usepackage[caption = false]{subfig}

\usepackage{subfig}

\newcommand{\mmkg}{\textsc{Mmkg}\xspace}

\makeatletter
\newcommand{\printfnsymbol}[1]{%
	\textsuperscript{\@fnsymbol{#1}}%
}
\makeatother

\title{MMKG: Multi-Modal Knowledge Graphs}

\author{Ye Liu\inst{1}\thanks{Contributed equally. Work done while at NEC Labs Europe.} \and
	Hui Li\inst{2}\printfnsymbol{1} \and
	Alberto Garcia-Duran\inst{3}\printfnsymbol{1} \and
	Mathias Niepert\inst{4} \and
	Daniel Onoro-Rubio\inst{4} \and
	David S. Rosenblum\inst{1}}

\authorrunning{Y. Liu et al.}

\institute{National University of Singapore, Singapore \\
	\email{\{liuye,david\}@comp.nus.edu.sg} \and
	Xiamen University, China \\
	\email{huili.xmu@gmail.com} \and
	EPFL, Switzerland \\
	\email{alberto.duran@epfl.ch} \and
	NEC Labs Europe, Germany \\
	\email{\{mathias.niepert, daniel.onoro\}@neclab.eu}}

\graphicspath{{figures/}}

\begin{document}
\maketitle
\noindent Github Repository: \url{https://github.com/nle-ml/mmkb}\\
\noindent Permanent URL: \url{https://zenodo.org/record/1245698}

\begin{abstract}
We present \mmkg, a collection of three knowledge graphs that contain both numerical features and (links to) images for all entities as well as entity alignments between pairs of KGs. Therefore, multi-relational link prediction and entity matching communities can benefit from this resource. We believe this data set has the potential to facilitate the development of novel multi-modal learning approaches for knowledge graphs. We validate the utility of \mmkg in the $\mathtt{sameAs}$ link prediction task with an extensive set of experiments. These experiments show that the task at hand benefits from learning of multiple feature types.
\end{abstract}

\section{Introduction}
\label{sec:intro}
A large volume of human knowledge can be represented with a multi-relational graph. Binary relationships encode \textit{facts} that can be represented in the form of RDF \cite{klyne2004resource} type triples $(\mathtt{head}, \mathtt{predicate}, \mathtt{tail})$, where $\mathtt{head}$ and $\mathtt{tail}$ are entities and $\mathtt{predicate}$ is the relation type. The combination of all triples form a multi-relational graph, where nodes represent entities and directed edges represent relationships. The resulting multi-relational graph is often referred to as a Knowledge Graph.

Knowledge Graphs (KGs) provide ways to efficiently organize, manage and retrieve this type of information, being increasingly used as external source of knowledge for problems like recommender systems~\cite{zhang2016collaborative}, language modeling~\cite{ahn2016neural}, question answering~\cite{YihCHG15} or image classification \cite{marino2016more}. While ranging from general purpose (\textsc{DBpedia}~\cite{AuerBKLCI07} or \textsc{Freebase}~\cite{BollackerEPST08}) to domain-specific (\textsc{IMDb} or \textsc{UniProtKB}), KGs are often highly incomplete and, therefore, research has focused heavily on the problem of knowledge graph completion~\cite{nickel2016review}. Link prediction (i.e. predicting missing relationships between the entities of the KG), relationship extraction~\cite{riedel2013relation} (i.e. classification of semantic relationship mentions) and ontology matching~\cite{SuchanekAS11} (i.e. alignment and integration of entities and relationships across KGs) are some of the different ways to tackle the incompleteness problem.

Novel data sets for benchmarking knowledge graph completion approaches, therefore, are important contributions to the community. This is especially true since one method performing well on one data set might perform poorly on others~\cite{toutanova2015observed}. With this paper we introduce \mmkg (Multi-Modal Knowledge Graphs), a collection of three knowledge graphs for link prediction and entity matching research. Contrary to existing data sets, these knowledge graphs contain both numerical features and images for all entities as well as entity alignments between pairs of KGs. There is a fundamental difference between \mmkg and  other visual-relational resources (e.g. \cite{krishnavisualgenome,wu2016ask}) . While  \mmkg is intended to perform relational reasoning across different entities and images, previous resources are intended to perform visual reasoning within the same image.

We use \textsc{Freebase15k}~\cite{bordes2013translating} as the blue print for the multi-modal knowledge graphs we constructed. \textsc{Freebase15k} is the major benchmark data set in the recent link prediction literature. In a first step, we aligned most FB15k entities to entities from \textsc{Dbpedia} and \textsc{Yago} through the $\mathtt{sameAs}$ links contained in \textsc{DBpedia} and \textsc{Yago} dumps. Since the degree of a node relates to the probability of an entity to appear in a subsampled version of a KG, we use this measure to populate our versions  of \textsc{DBpedia} and \textsc{Yago} with more entities. For each knowledge graph, we  include entities that are highly connected to the aligned entities so that the number of entities in each KG is similar to that of \textsc{FB15k}.  Lastly, we have populated the three knowledge graphs with numeric literals and images for (almost) all of their entities. We name the two new data sets \textsc{Dbpedia15k} and \textsc{Yago15k}. Although all three data sets contain a similar number of entities, this does not prevent potential users of \mmkg from filtering out entities to benchmark  approaches in scenarios where KGs largely differ with respect to the number of entities  that they contain.

The contributions of the present paper are the following:
\begin{itemize}
\item The creation of two knowledge graphs \textsc{DBpedia15k} and \textsc{YAGO15k}, that are the \textsc{DBpedia} and \textsc{YAGO}~\cite{suchanek2007yago} counterparts, respectively, of \textsc{Freebase15k}. Furthermore, all three KGs are enriched with numeric literals and image information, as well as $\mathtt{sameAs}$ predicates linking entities from pairs of knowledge graphs. $\mathtt{sameAs}$ predicates, numerical literals and (links to) images for entities so as the relational graph structure are released in separate files.
\item We validate our hypothesis that knowledge graph completion related problems can benefit from multi-modal data:
\begin{itemize}
\item We elaborate on a previous learning framework~\cite{garcia2017kblrn} and extend it by also incorporating image information. We perform completion in queries such as $(\mathtt{head?}, \mathtt{sameAs}, \mathtt{tail})$ and $(\mathtt{head}, \mathtt{sameAs}, \mathtt{tail?})$, where $\mathtt{head}$ and $\mathtt{tail}$ are entities, each one from a different KG. This task can be deemed something in-between link prediction and entity matching.
\item We analyze the performance of the different modalities in isolation for different percentages of known aligned entities between KGs, as well as for different combinations of feature types.
\end{itemize}
\end{itemize}

The paper is organized as follows: In Section~\ref{sec:rel} we discuss the relevance of \mmkg for link prediction and entity matching research. Section~\ref{sec:data} elaborates on how the different elements of \mmkg were constructed and provides relevant statistics of the resource. Section~\ref{sec:methods} presents the learning framework and our extension, which is followed by experimental evidence in Section~\ref{sec:experiments} that validates our hypothesis about the need of such data set. Finally, Section~\ref{sec:conclusion} presents our conclusions.

\begin{figure*}[t!]
\centering
\includegraphics[width=1.0\textwidth]{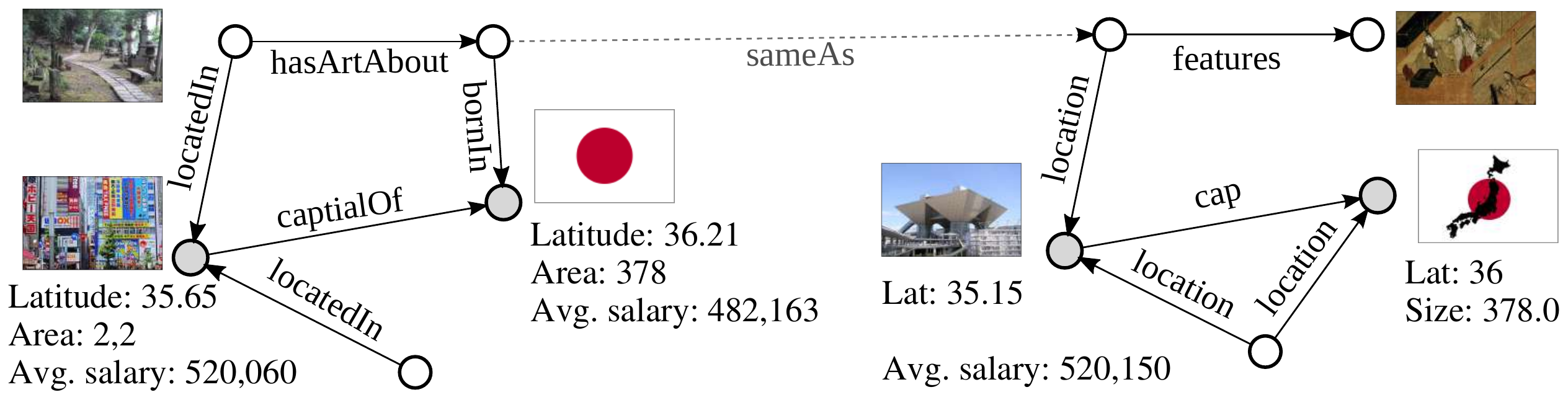}
\caption{\label{fig:KG-example-and-queries} Illustration of \mmkg.}
\end{figure*}

\section{Relevance}
\label{sec:rel}

There are a number of problems related to knowledge graph completion. \textit{Named-entity linking} (NEL)~\cite{dredze2010entity,hajishirzi2013joint} is the task of linking a named-entity mention from a text to an entity in a knowledge graph. Usually a NEL algorithm is followed by a second procedure, namely \textit{relationship extraction}~\cite{mintz2009distant,riedel2013relation}, which aims at linking relation mentions from text to a canonical relation type in a knowledge graph. Hence, relation extraction methods are often used in conjunction with NEL algorithms to perform KG completion from natural language content.

Link prediction and entity matching are two other popular tasks for knowledge graph completion. \mmkg has been mainly created targeting these two tasks.

\textbf{Link prediction.} It aims at answering completion queries of the form $(\mathtt{head?},$ $\mathtt{predicate},$ $\mathtt{tail})$ or $(\mathtt{head}, \mathtt{predicate}, \mathtt{tail?})$, where the answer is supposed to be always within the KG.

\textbf{Entity Matching.} Given two KGs, the goal is to find pairs of records, one from each KG, that refer to the same entity. For instance, $\mathtt{DBpedia}$:$\mathtt{NYC} \equiv \mathtt{FB}$:$\mathtt{New York}$.

\subsection{Relevance for Multi-Relational Link Prediction Research}

The core of most of multi-relational link prediction approaches is a scoring function. The scoring function is a (differentiable) function whose parameters are learned such that it assigns high scores to true triples and low scores to triples assumed to be false. The majority of recent work fall into one of the following two categories:
\begin{enumerate}
\item Relational approaches~\cite{lao2011random,gardner2015efficient} wherein features are given as logical formulas which are evaluated in the KG to determine the feature's value. For instance, the formula $\exists x\ (\mathtt{A}, \mathtt{bornIn}, x) \wedge (x, \mathtt{capitalOf}, \mathtt{B})$ corresponds to a binary feature which is $1$ if there exists a path of that type from entity $\mathtt{A}$ to entity $\mathtt{B}$, and $0$ otherwise.
\item Latent approaches~\cite{nickel2016review} learn fixed-size vector representations (embeddings) for all entities and relationships in the KG.
\end{enumerate}

While previous work has almost exclusively focused on the relational structure of the graph, recent approaches have considered other feature types like numerical literals~\cite{garcia2017kblrn,Pezeshkpour2018EmbeddingMR}. In addition, recent work on visual-relational knowledge graphs~\cite{onoro2017representation} has introduced novel visual query types such as "\textit{How are these two unseen images related to each other?}" and has proposed novel machine learning methods to answer these queries. Different to the link prediction problem addressed in this work, the methods evaluated in \cite{onoro2017representation} solely rely on visual data.

\mmkg provides three data sets for evaluating multi-relational link prediction approaches where, in addition to the multi-relational links between entities, all entities have been associated with numerical and visual data. An interesting property of \mmkg is that the three knowledge graphs are very heterogeneous (w.r.t. the number of relation types, their sparsity, and so on) as we show in Section~\ref{sec:data}.  It is known that the performance of multi-relational link prediction methods depends on the characteristics of the specific knowledge graphs~\cite{toutanova2015observed}. Therefore, \mmkg is an important benchmark data set for measuring the robustness of the approaches.

\subsection{Relevance for Entity Matching Research}

There are numerous approaches to find $\mathtt{sameAs}$ links between entities of two different knowledge graphs. Though there are works~\cite{niu2012effective,galarraga2013mining} that solely incorporate the relational graph structure, there is an extensive literature on methods that perform the matching by combining relational structural information with literals of entities, where literals are used to compute prior confidence scores~\cite{suchanek2011paris,lacoste2013sigma,noessner2010leveraging}.

A large number of approaches of the entity matching literature have been evaluated as part of the Ontology Alignment Evaluation Initiative (OAEI) \cite{achichi2016results} using data sets such as \textsc{Yago}, \textsc{Freebase}, and \textsc{Imdb}\cite{lacoste2013sigma,suchanek2011paris,noessner2010leveraging}. Contrary to the proposed multi-modal knowledge graph data sets, however, the OAEI does not focus on tasks with visual and numerical data.
The main advantages of \mmkg over existing benchmark data sets for entity matching are: (1) \mmkg's entities are associated with visual and numerical data, and (2) the availability of ground truth entity alignments for a high percentage of the KG entities. The former encourages research in entity matching methods that incorporate visual and numerical data. The latter allows one to measure the robustness in performance of entity matching approaches with respect to the number of given alignments between two KGs. The benchmark KGs can also be used to evaluate different active learning strategies. Traditional active learning approaches ask a user for a small set of alignments that minimize the uncertainty and, therefore, maximize the quality of the final alignments.

\begin{table}[t!]
	\caption{Files from which we extract the different subcomponents of \mmkg.}
	\label{table:links}
	\resizebox{\textwidth}{!}{
\begin{tabular}{|c|l|}
  \hline
  \multicolumn{2}{|c|}{\textsc{DB15k}}\\ \hline
  $\mathtt{sameAs}$ & {\small http://downloads.dbpedia.org/2016-10/core-i18n/en/freebase\_links\_en.ttl.bz2} \\ \hline
  \multirow{2}{*}{Relational Graph} & {\small http://downloads.dbpedia.org/2016-10/core-i18n/en/mappingbased\_objects\_en.ttl.bz2} \\
  & {\small http://downloads.dbpedia.org/2016-10/core-i18n/en/instance\_types\_en.ttl.bz2} \\  \hline
  \multirow{3}{*}{Numeric Literals} & {\small http://downloads.dbpedia.org/2016-10/core-i18n/en/geo\_coordinates\_en.tql.bz2} \\
  & {\small http://downloads.dbpedia.org/2016-10/core-i18n/en/mappingbased\_literals\_en.tql.bz2}  \\
  & {\small http://downloads.dbpedia.org/2016-10/core-i18n/en/persondata\_en.tql.bz2} \\
  \hline \hline
    \multicolumn{2}{|c|}{\textsc{Yago15k}}\\ \hline
  $\mathtt{sameAs}$ & {\small http://resources.mpi-inf.mpg.de/yago-naga/yago3.1/yagoDBpediaInstances.ttl.7z} \\ \hline
  Relational Graph & {\small http://resources.mpi-inf.mpg.de/yago-naga/yago3.1/yagoFacts.ttl.7z} \\ \hline
  \multirow{2}{*}{Numeric Literals} & {\small http://resources.mpi-inf.mpg.de/yago-naga/yago3.1/yagoDateFacts.ttl.7z} \\
  & {\small http://resources.mpi-inf.mpg.de/yago-naga/yago3.1/yagoGeonamesOnlyData.ttl.7z}  \\
  \hline
\end{tabular}}
\end{table}

\section{\mmkg: Dataset Generation}
\label{sec:data}

We chose \textsc{Freebase-15k} (\textsc{FB15k}), a data set that has been widely used in the knowledge graph completion literature, as a starting point to create the multi-modal knowledge graphs. Facts of this KG are in N-Triples format, a line-based plain text format for encoding an RDF graph. For example, the triple
\begin{center}
{\small \texttt{</ns/g.112ygbz6>    </ns/type.object.type>   </ns/film.film>}}.
\end{center}
indicates that the entity with identifier \texttt{</ns/g.112ygbz6>} is connected to the entity with identifier \texttt{</ns/film.film>} via the relationship \texttt{</ns/type.object.type>}.\\


\begin{table}[t!]
\centering
\caption{Statistics of the \mmkg knowledge graphs.}
\label{data-statistics}
\begin{tabular}{lcc|cccc|}
\cline{4-7}
                              & \multicolumn{1}{l}{}           & \multicolumn{1}{l|}{}                & \multicolumn{4}{c|}{Number of Triples}                                                                                                 \\ \hline
\multicolumn{1}{|l|}{KG}      & \multicolumn{1}{l}{\#Entities} & \multicolumn{1}{l|}{\#Relationships} & \multicolumn{1}{l}{\ \ Relational Graph \ \ } & \multicolumn{1}{l}{\ \ Numeric Literals \ \ } & \multicolumn{1}{l}{\ \ Images \ \ } & \multicolumn{1}{l|}{sameAs} \\ \hline
\multicolumn{1}{|l|}{FB15k}   & 14,951                         & 1,345                                & 592,213                              & 29,395                               & 13,444                     & -                           \\
\multicolumn{1}{|l|}{DB15k}   & 14,777                         & 279                                  & 99,028                               & 46,121                               & 12,841                     & 12,846                      \\
\multicolumn{1}{|l|}{Yago15k} & 15,283                         & 32                                   & 122,886                              & 48,405                               & 11,194                     & 11,199                      \\ \hline
\end{tabular}
\end{table}


We create versions of \textsc{DBpedia} and \textsc{Yago}, called \textsc{DBpedia-15k} (\textsc{DB15k}) and \textsc{Yago15k}, by aligning entities in \textsc{FB15k} with entities in these other knowledge graphs. More concretely, for \textsc{DB15k} we performed the following steps.

\begin{enumerate}
\item \textsc{\textbf{sameAs.}} We extract alignments between entities of \textsc{FB15k} and \textsc{DBpedia} in order to create \textsc{DB15k}. These alignments link one entity from \textsc{FB15k} to one from \textsc{DBpedia} via a $\mathtt{sameAs}$ relation.

\item \textsc{\textbf{Relational Graph.}} A high percentage of entities from \textsc{FB15k} can be aligned with entities in  \textsc{DBpedia}. However, to make the two knowledge graphs have roughly the same number of entities and to also have entities that cannot be aligned across the knowledge graphs, we include additional entities in \textsc{DB15k}.  We chose entities with the highest connectivity to the already aligned entities to complete \textsc{DB15k}. We then collect all the triples where both $\mathtt{head}$ and $\mathtt{tail}$ entities belong to the set of entities of \textsc{DB15k}. This collection of triples forms the relational graph structure of \textsc{DB15k}.

\item \textsc{\textbf{Numeric Literals.}} We collect all triples that associate entities in \textsc{DB15k} with numerical literals. For example, the relations \texttt{/location/geocode/latitude} links entities to their latitude. We refer to these relation types as numerical relations. Figure~\ref{fig:num-rel} shows the most common numerical relationships in the knowledge graphs. In previous work~\cite{garcia2017kblrn} we have extracted numeric literals for \textsc{FB15k} only.

\begin{figure}[h!]
\begin{minipage}{.5\linewidth}
\centering
\includegraphics[width=0.9\textwidth]{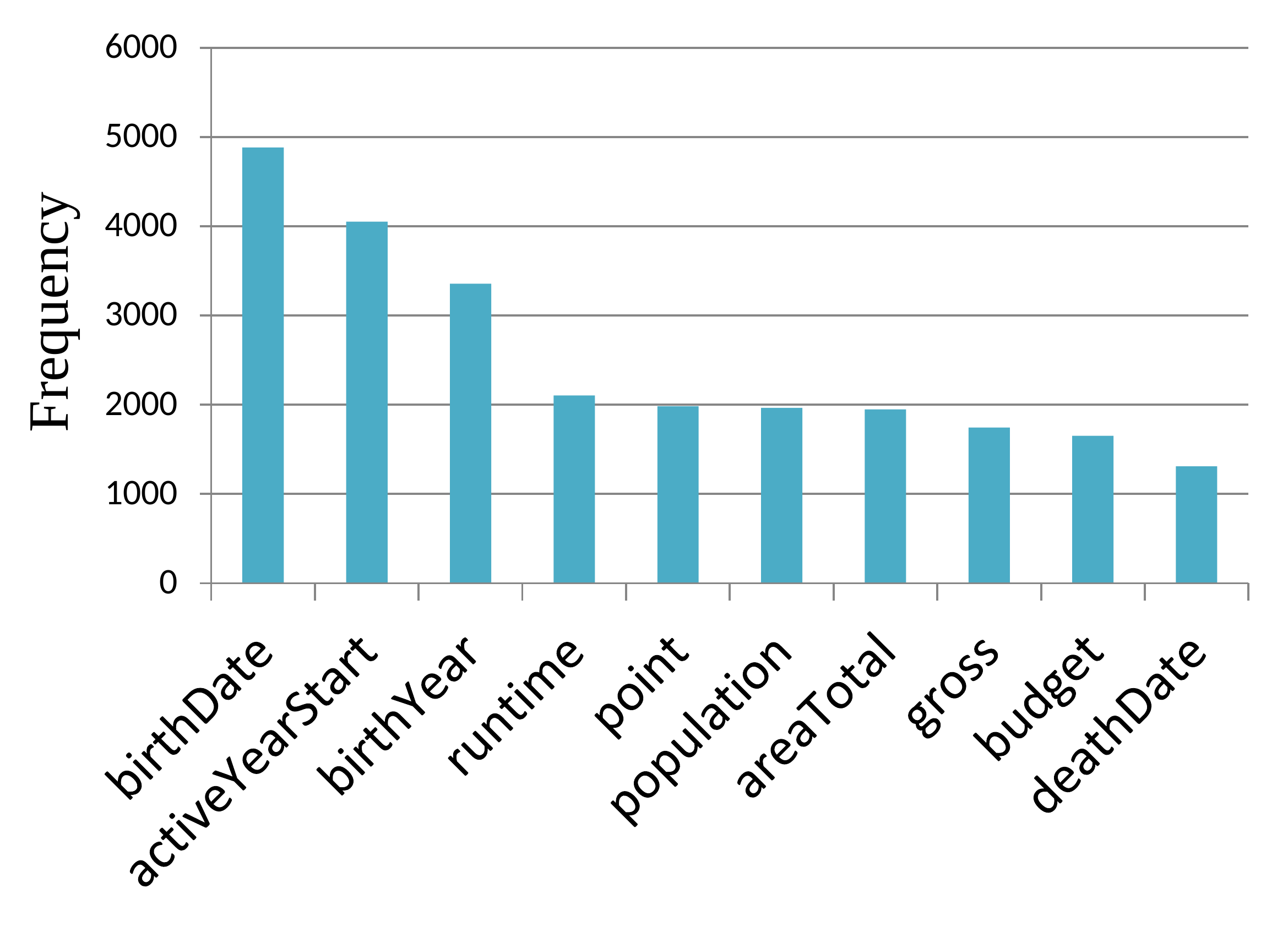}
\end{minipage}%
\begin{minipage}{.5\linewidth}
\centering
\includegraphics[width=0.92\textwidth]{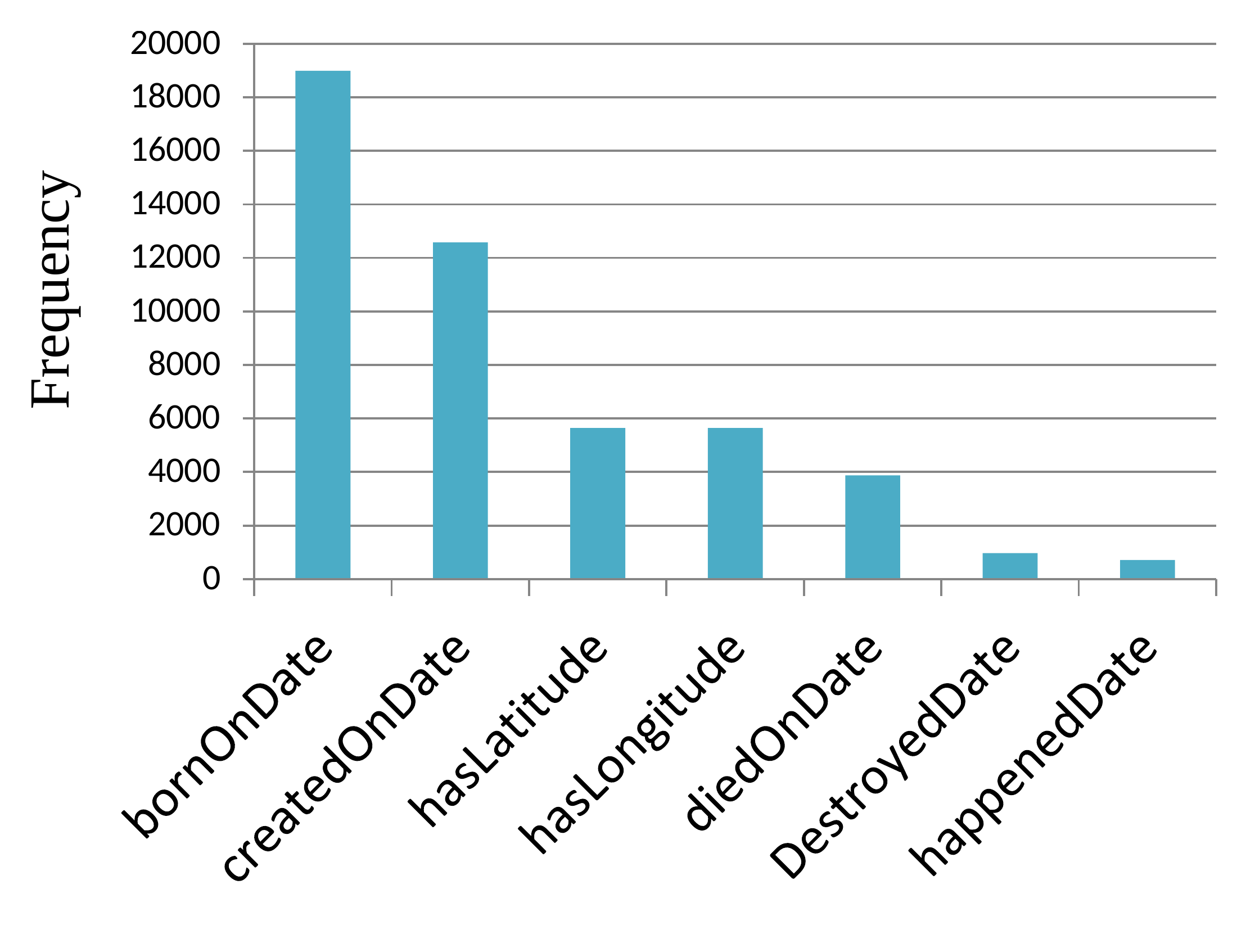}
\end{minipage}
\caption{Most common numerical relationships in \textsc{DB15k} (left) and \textsc{Yago15k} (right). }
\label{fig:num-rel}
\end{figure}

\item \textsc{\textbf{Images.}} We obtain images related to each of the entities of \textsc{FB15k}. To do so we implemented a web crawler that is able to parse query results for the image search engines Google Images, Bing Images, and Yahoo Image Search. To minimize the amount of noise due to polysemous entity labels (for example, there are two \textsc{Freebase} entities with the text label ``Paris") we extracted, for each entity in \textsc{FB15k}, all Wikipedia URIs from the $1.9$ billion triple \textsc{Freebase} RDF dump\footnote{\url{https://developers.google.com/freebase/}}. For instance, for Paris, we obtained URIs such as \texttt{Paris(ile-de-France,France)} and \texttt{Paris(City\_of\_New\_Orleans, Louisiana)}. These URIs were processed and used as search queries for  disambiguation purposes. We crawled web images also following other type of search queries, and not only the  Wikipedia URIs. For  example, we used i) the entity name, and ii) the entity name followed by the entity's notable type as query  strings, among others. After visual inspection of polysemous entities (as they are the most problematic entities), we observed that using Wikipedia  URIs as query strings was the strategy that alleviated most the polysemy problem.
We used the crawler to download a number of images per entity. For each entity we stored the 20 top ranked images retrieved by each browser.  We filtered out  images with a side  smaller than 224 pixels, and images with a side 2.5 bigger than the other. We also removed corrupted, low quality, and duplicate images (pairs of images with a pixel-wise distance  below a certain threshold). After all these  steps, we kept 55.8 images per entity on average. We also scaled the images to have a maximum height or width of 500 pixels while maintaining their aspect ratio. Finally, for each entity we distribute a distinct image to \textsc{FB15k} and \textsc{DB15k}.

\end{enumerate}

We repeat the same sequence of steps for the creation of \textsc{Yago15k} with one difference. $\mathtt{sameAs}$ predicates from the \textsc{Yago} dump align entities from that knowledge graph to \textsc{DBpedia} entities. We used them along with the previously extracted alignments between \textsc{DB15k} and \textsc{FB15k} to eventually create the alignment between \textsc{Yago} and \textsc{FB15k} entities.
Table \ref{table:links} depicts the hyperlinks from where we extracted the different component for the generation of \textsc{DB15k} and \textsc{Yago15k}.

\begin{figure}[t!]
\begin{minipage}{.5\linewidth}
\centering
\includegraphics[width=1\textwidth]{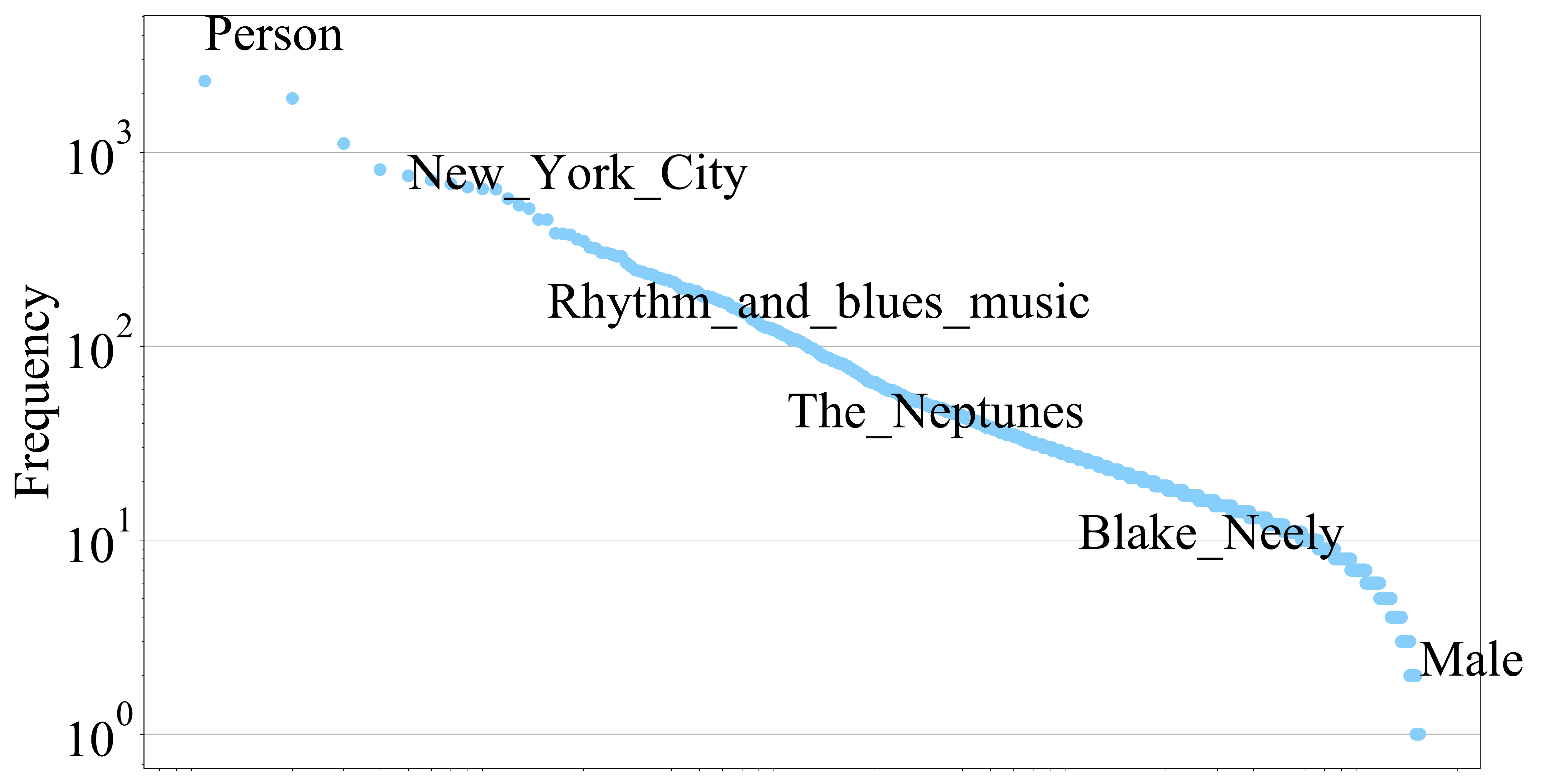}
\end{minipage}%
\begin{minipage}{.5\linewidth}
\centering
\includegraphics[width=1\textwidth]{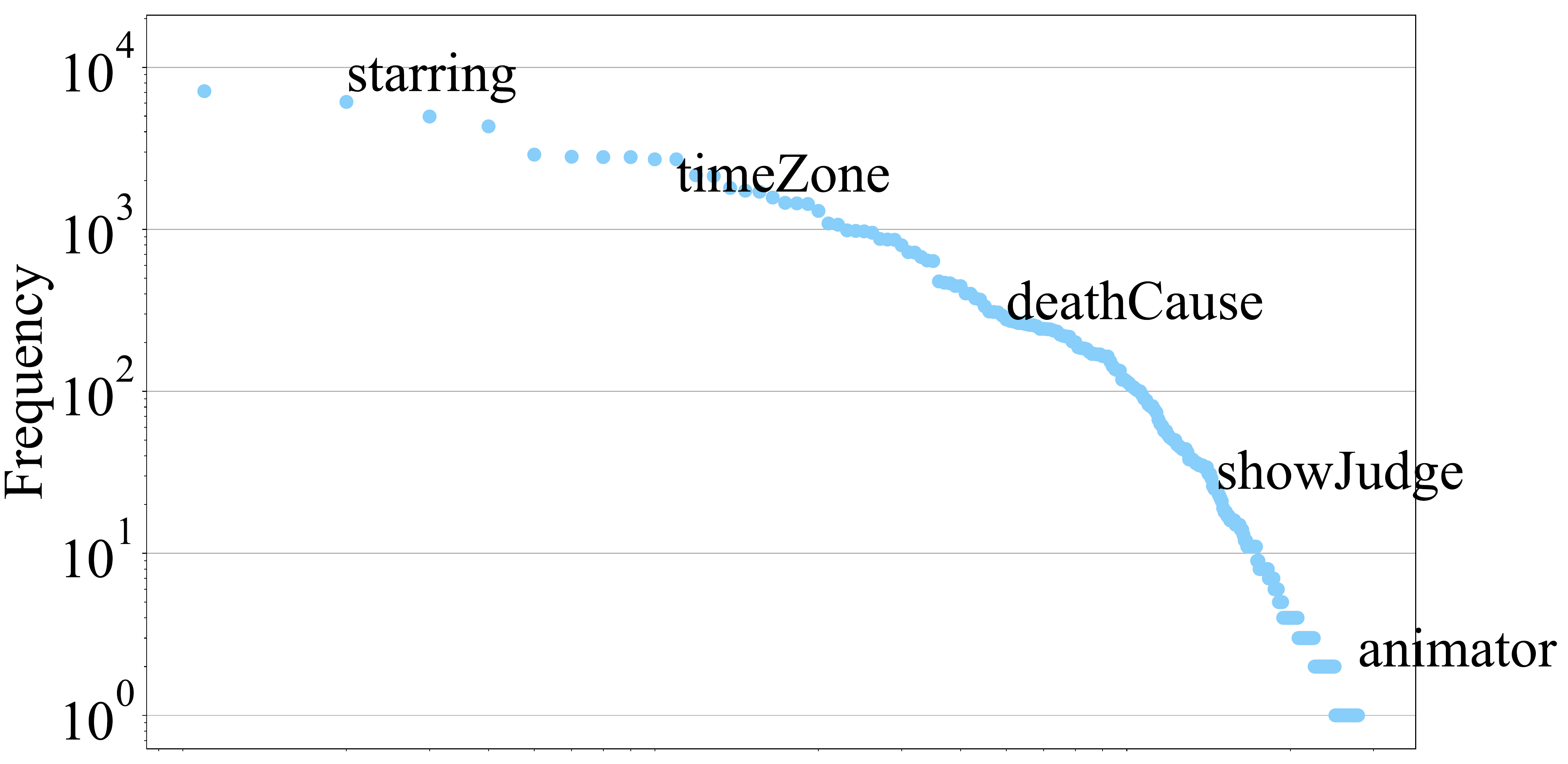}
\end{minipage}
\caption{ Entity (left) and relation type (right) frequencies in \textsc{Yago15k}.}
\label{fig:rel-freq:Yago15k}
\end{figure}

\begin{figure}[t!]
\begin{minipage}{.5\linewidth}
\centering
\includegraphics[width=1\textwidth]{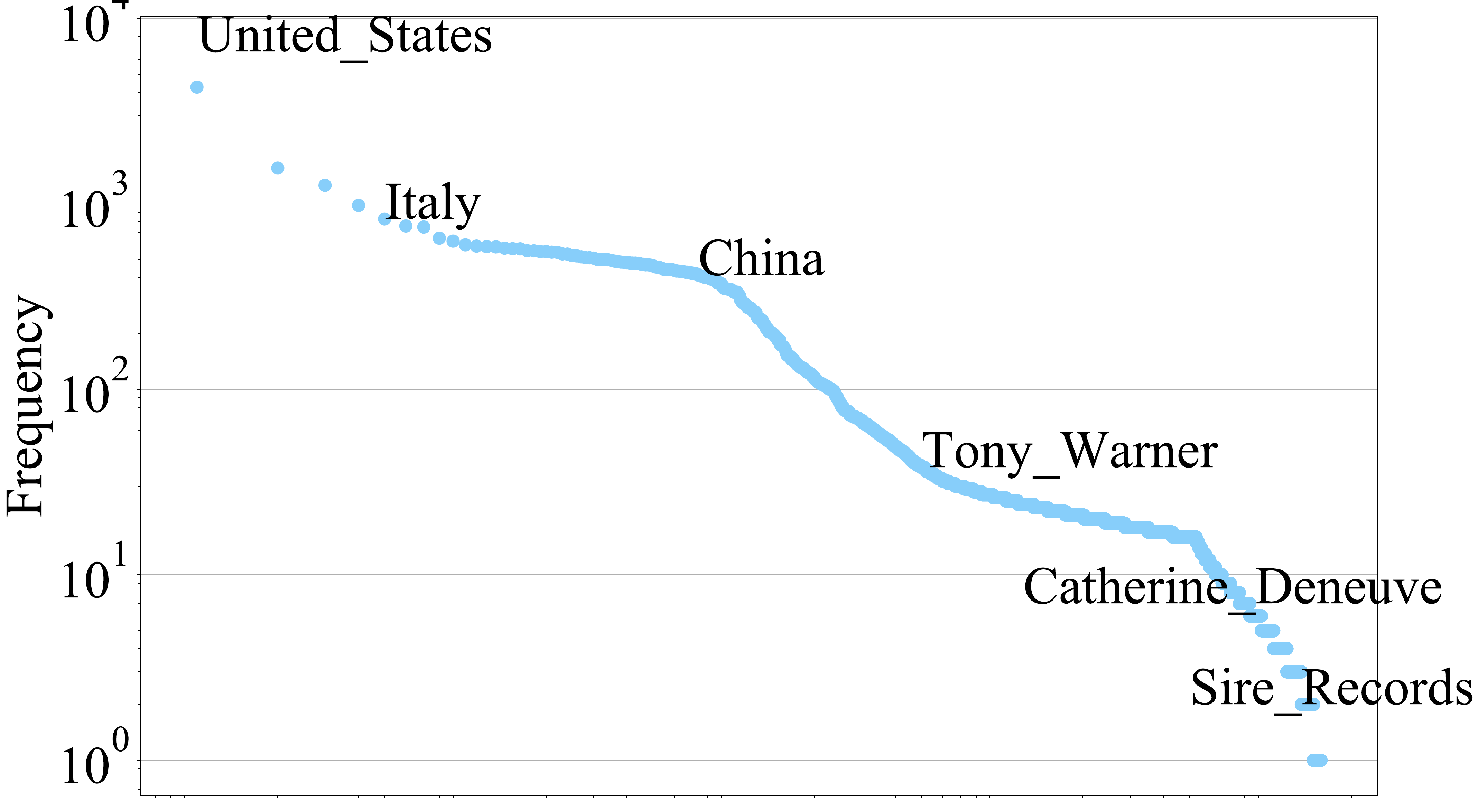}
\end{minipage}%
\begin{minipage}{.5\linewidth}
\centering
\includegraphics[width=1\textwidth]{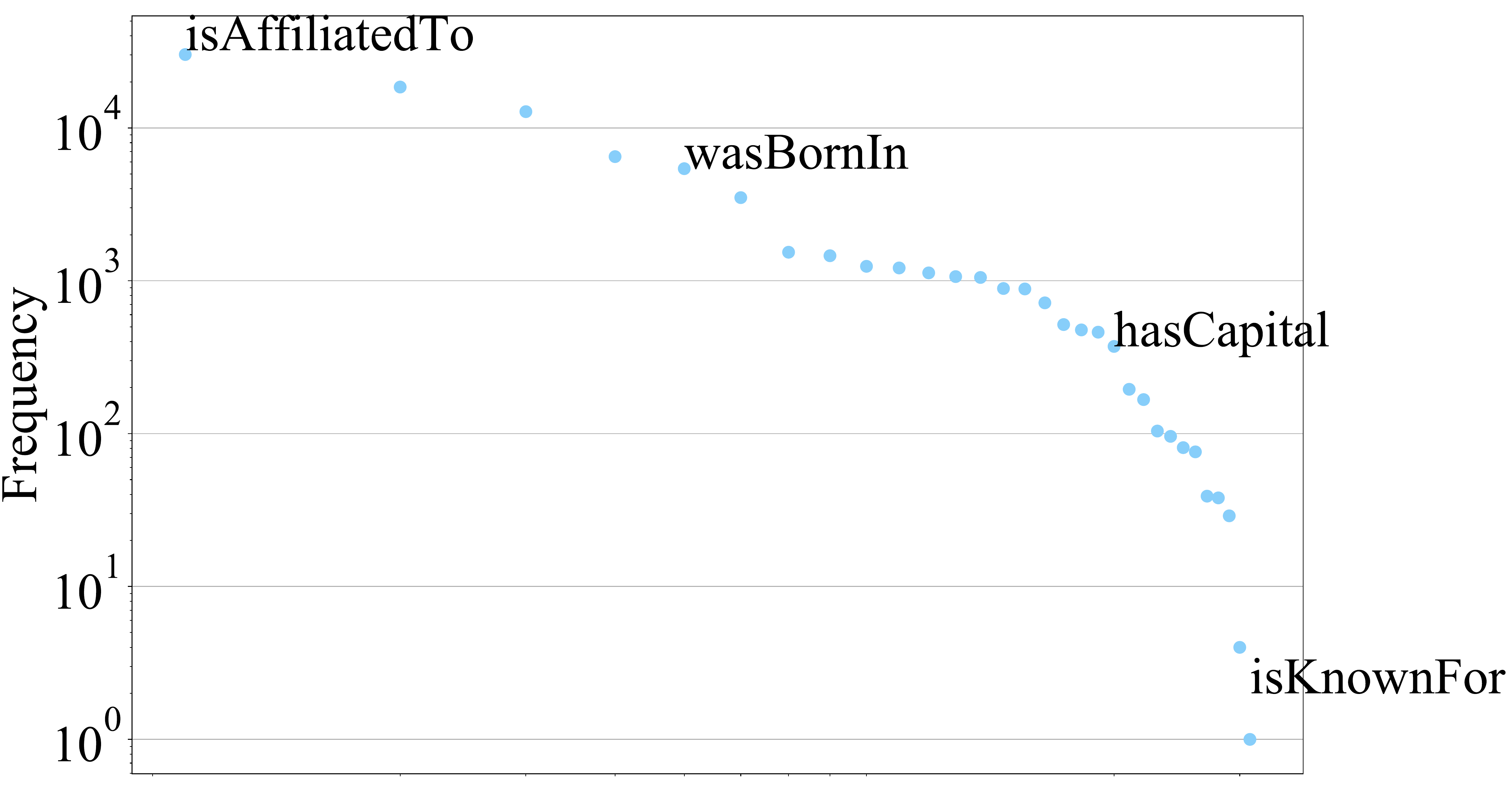}
\end{minipage}
\caption{ Entity (left) and relation type (right) frequencies in \textsc{DB15k}.}
\label{fig:rel-freq:DB15k}
\end{figure}

Statistics of \textsc{FB15K}, \textsc{DB15K} and \textsc{Yago15k} are depicted in Table~\ref{data-statistics}. The frequency of entities and relationships in \textsc{Yago15k} and \textsc{DB15k} are depicted in Figure \ref{fig:rel-freq:Yago15k} and \ref{fig:rel-freq:DB15k}, respectively. Entities and relationships are sorted according to their frequency. They show in logarithmic scale the number of times that each entity and relationship occurs in \textsc{Yago15k} and \textsc{DB15k}. Relationships like $\mathtt{starring}$ or $\mathtt{timeZone}$ occur quite frequently in \textsc{Yago15k}, while others like $\mathtt{animator}$ are rare. Contrary to \textsc{Fb15k}, the entity $\mathtt{Male}$ is unusual in \textsc{Yago15k}, which illustrates, to a limited extent, the heterogeneity of the KGs.

\subsection{Availability and Sustainability}

\mmkg can be found in the Github repository \url{https://github.com/nle-ml}. We will actively use Github issues to track feature requests and bug reports. The documentation of the framework has been published on the repository's Wiki as well. To guarantee the future availability of the resource, it has also been published on Zenodo. \mmkg is released under the BSD-3-Clause License.

The repository contains a number of files, all of them formatted following the N-Triples guidelines (\url{https://www.w3.org/TR/n-triples/}). These files contain information regarding the relational graph structure, numeric literals and visual information. Numerical information is formatted as RDF literals, entities and relationships point to their corresponding RDF URIs\footnote{Unfortunately, Freebase has deprecated RDF URIs.}. We also provide separates files that link both \textsc{DB15k} and \textsc{Yago15k} entities to \textsc{FB15k} ones via $\mathtt{sameAs}$ predicates, also formatted as N-Triples.

\begin{figure*}[t!]
\centering
\includegraphics[width=0.9\textwidth]{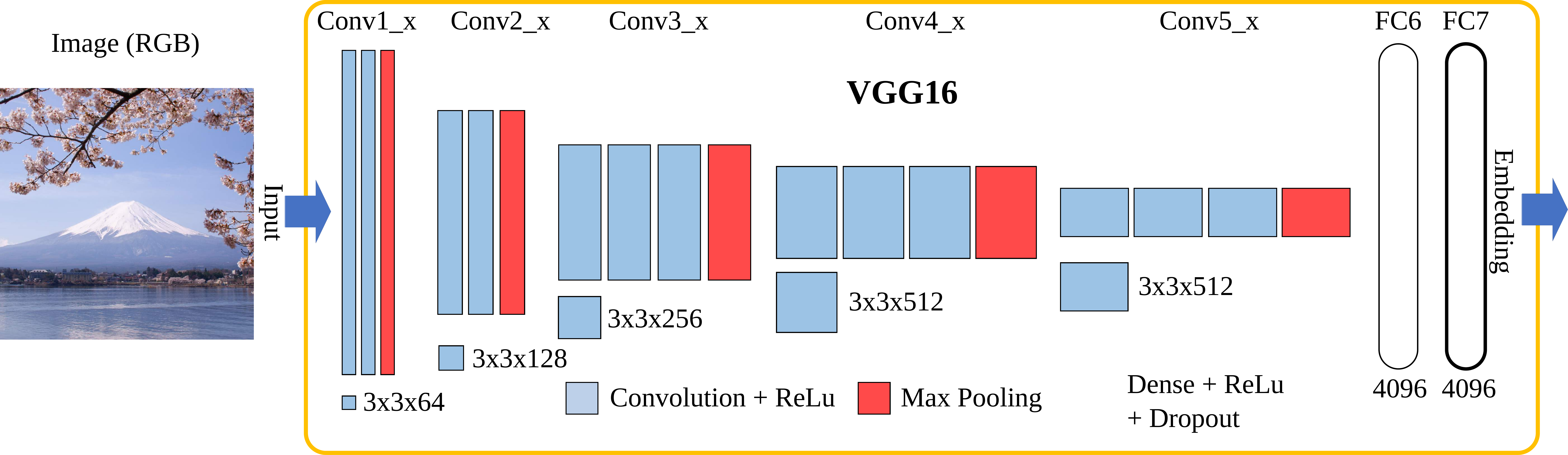}
\caption{\label{fig:VGG} Low-dimensional embeddings learned for images through VGG16.}
\end{figure*}

To avoid copyright infringement and guarantee the access to the visual information (i.e. URLs to images are not permanent), we learn embeddings for the images through the VGG16 model introduced in \cite{simonyan2014very}. The VGG16 model used for this work was previously trained on the ILSVRC 2012 data set derived from \textsc{ImageNet}~\cite{deng2009imagenet}. The architecture of this network is illustrated in Figure \ref{fig:VGG}. We remove the softmax layer of the trained VGG16 and obtain the 4096-dimensional embeddings for all images of \mmkg. We provide these embeddings in hdf5~\cite{hdf5} format. The Github repository contains documentation on how to access these embeddings. Alternatively, one can use the crawler (also available in the Github repository) to download the images from the different search engines.

\section{Technical Quality of \mmkg}
\label{sec:methods}

We provide empirical evidence that knowledge graph completion related tasks can benefit from the multi-modal data of \mmkg. Our hypothesis is that different data modalities contain complementary information beneficial for both multi-relational link prediction and entity matching. For instance, in the entity matching problem if two images are visually similar they are likely to be associated with the same entity and if two entities in two different KGs have similar numerical feature values, they are more likely to be identical. Similarly, we hypothesize that multi-relational link prediction can benefit from the different data modalities. For example, learning that the mean difference of birth years is 0.4 for the Freebase relation $\mathtt{/people/marriage/spouse}$, can provide helpful evidence for the linking task.


In recent years, numerous methods for merging feature types have been proposed. The most common strategy is the concatenation of either the input features or some intermediate learned representation. We compare these strategies to the recently proposed learning framework~\cite{garcia2017kblrn}, which we have found to be superior to the concatenation and an ensemble type of approach.

\begin{figure*}[t!]
\centering
\includegraphics[width=\textwidth]{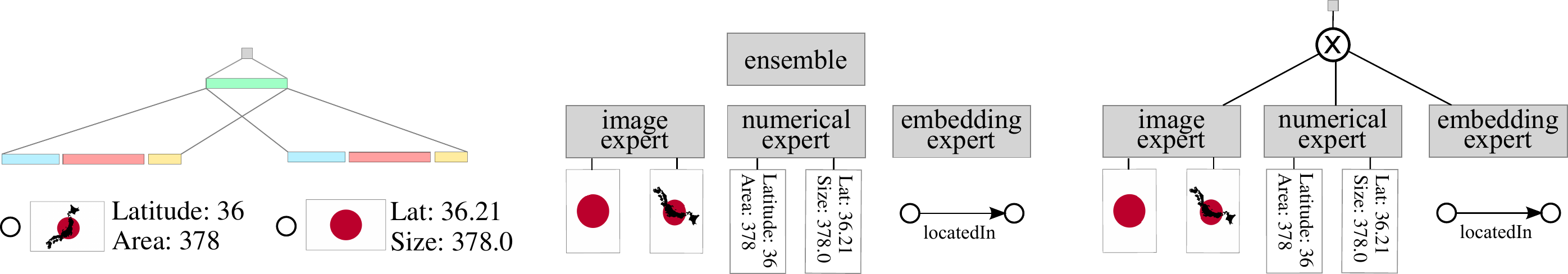}
\caption{\label{fig:KG-example-and-queries} Illustration of the methods we evaluated to combine various data modalities.}
\end{figure*}

\subsection{Task: \textsc{sameAs} Link Prediction}

We validate the hypothesis that different modalities are complementary for the $\mathtt{sameAs}$ link prediction task.
Different to the standard link prediction problem, here the goal is to answer queries such as $(\mathtt{head?}, \mathtt{sameAs}, \mathtt{tail})$ or $(\mathtt{head}, \mathtt{sameAs}, \mathtt{tail?})$ where $\mathtt{head}$ and $\mathtt{tail}$ are entities from different KGs. We do not make the one-to-one alignment assumption, that is, the assumption that one entity in one KG is identical to exactly (at most) one in the other. A second difference is that in the evaluation of the \textsc{SameAs} prediction task, and in general in the link prediction literature, \textit{only} one argument of a triple is assumed to be missing at a time. That partial knowledge of the ground truth is not given in the entity matching literature.

\subsection{Model: Products of Experts}
\label{sec:poe}
We elaborate on previous work~\cite{garcia2017kblrn} and extend it by incorporating visual information.
Such learning framework can be stated as a Product of Experts (PoE).


In general, a PoE's  probability distribution is
$$p( \mathbf{d} \mid \theta_1, ..., \theta_n) = \frac{\prod_i f_i(\mathbf{d} \mid \theta_i)}{\sum_{\mathbf{c}} \prod_i f_i(\mathbf{c} \mid \theta_{i})},$$
where $\mathbf{d}$ is a data vector in a discrete space, $\theta_i$ are the parameters of individual model $f_i$, $f_i( \mathbf{d} \mid \phi_i)$ is the value of $d$ under model $f_i$, and the $\mathbf{c}$'s index all possible vectors in the data space. The PoE model is now trained to assign high probability to observed data vectors.

In the KG context, the data vector $\mathbf{d}$ is always a triple $\mathtt{d} = (\mathtt{h}, \mathtt{r}, \mathtt{t})$ and the objective is to learn a PoE that assigns high probability to true triples and low probabilities to triples assumed to be false. For instance, the triple $(\mathtt{Paris}, \mathtt{locatedIn}, \mathtt{France})$ should be assigned a high probability and the triple $(\mathtt{Paris}, \mathtt{locatedIn}, \mathtt{Germany})$ a low probability. If $(\mathtt{h},\mathtt{r},\mathtt{t})$ holds in the KG, the pair's vector representations are used as  positive training examples.
Let $\mathtt{d} = (\mathtt{h},\mathtt{r}, \mathtt{t})$. We can now define one individual expert $f_{(\mathtt{r},\mathtt{F})}(\mathtt{d} \mid \phi_{(\mathtt{r},\mathtt{F})})$ for each (relation type $\mathtt{r}$, feature type $\mathtt{F}$)  pair
\begin{equation*}
\label{eq:KB-C}
\begin{split}
f_{(\mathtt{r},\mathtt{L})}(\mathtt{d} \mid \theta_{(\mathtt{r},\mathtt{L})}): & \mbox{ the embedding expert for relation type } \mathtt{r}    \\
f_{(\mathtt{r},\mathtt{R})}(\mathtt{d} \mid \theta_{(\mathtt{r},\mathtt{R})}): & \mbox{ the relational expert for relation type } \mathtt{r}  \\
f_{(\mathtt{r},\mathtt{N})}(\mathtt{d} \mid \theta_{(\mathtt{r},\mathtt{N})}): & \mbox{ the numerical expert for relation type } \mathtt{r}  \\
f_{(\mathtt{sameAs},\mathtt{I})}(\mathtt{d} \mid \theta_{(\mathtt{r},\mathtt{I})}): & \mbox{ the visual expert for relation type } \mathtt{sameAs} \\
\end{split}
\end{equation*}

The joint probability for a triple $\mathtt{d} = (\mathtt{h}, \mathtt{r}, \mathtt{t})$  of the PoE model is now
$$p( \mathtt{d} \mid \theta_1, ..., \theta_n) = \frac{\prod_{\mathtt{F} \in \{\mathtt{R, L, N, I}\}} f_{(\mathtt{r}, \mathtt{F})}(\mathtt{d} \mid \theta_{(\mathtt{r},\mathtt{F})})}{\sum_{\mathtt{c}} \prod_{\mathtt{F} \in \{\mathtt{R, L, N, I}\}} f_{(\mathtt{r}, \mathtt{F})}(\mathtt{c} \mid \theta_{(\mathtt{r},\mathtt{F})}))},$$
where $\mathtt{c}$ indexes all possible triples.  


For information regarding the latent, relational and numerical experts, we refer the reader to \cite{garcia2017kblrn}. Although entity names are not used to infer $\mathtt{sameAs}$ links in this work, one may also define an expert for such feature.

\subsubsection{Visual Experts}
The visual expert is only learned for the $\mathtt{sameAs}$ relation type. The scores for the image experts is computed by the cosine similarity between two 4096-dimensional feature vectors from the two images.

Let $\mathtt{d} = (\mathtt{h}, \mathtt{r}, \mathtt{t})$ be a triple. The visual expert for relation type $\mathbf{r}$ is defined as
\begin{equation*}
\label{eq:KB-C}
\begin{split}
f_{(\mathtt{r},\mathtt{I})}(\mathtt{d} \mid \theta_{(\mathtt{r},\mathtt{I})}) = & \ \exp\left(\mathbf{i}_{\mathtt{h}} \cdot \mathbf{i}_{\mathtt{t}}\right)  \mbox{ and } \\
f_{(\mathtt{r'},\mathtt{I})}(\mathtt{d} \mid \theta_{(\mathtt{r'},\mathtt{I})}) = & \ 1 \mbox { for all } \mathtt{r'} \neq \mathtt{r} ,
\end{split}
\end{equation*}
where $\cdot$ is the dot product and $\mathbf{i}_{\mathtt{h}}$ and $\mathbf{i}_{\mathtt{t}}$ are embeddings of the images for the $\mathtt{head}$ and $\mathtt{tail}$ entities.

\subsubsection{Learning}
The logarithmic loss for the given training triples $\mathbf{T}$ is defined as
$$\mathcal{L} = -\sum_{\mathtt{t} \in \mathbf{T}} \log p(\mathtt{t} \mid \theta_1, ..., \theta_n).$$
 To fit the PoE to the training triples, we follow the derivative of the log likelihood of each observed triple $\mathtt{d}\in \mathbf{T}$ under the PoE
\begin{equation*}
\begin{split}
\frac{\partial \log p( \mathtt{d} \mid \theta_1, ..., \theta_n) }{ \partial \theta_m } = & \frac{\partial \log f_i(\mathtt{d} \mid \theta_i)}{\partial \theta_m}
- \frac{\partial \log \sum_{\mathtt{c}} \prod_i f_i(\mathtt{c} \mid \theta_{i})}{\partial \theta_m}
\end{split}
\end{equation*}
We follow~\cite{garcia2017kblrn} and we generate for each triple $\mathtt{d} = (\mathtt{h}, \mathtt{r}, \mathtt{t})$ a set $\mathbf{E}$ consisting of $N$ triples $(\mathtt{h}, \mathtt{r}, \mathtt{t'})$  by sampling  exactly $N$ entities $\mathtt{t'}$ uniformly at random from the set of all entities. In doing so, the right term is then approximated by
$$\frac{\partial \log \sum_{\mathtt{c} \in \mathbf{E}} \prod_i f_i(\mathtt{c} \mid \theta_{i})}{\partial \theta_m}.$$ This is often referred to as negative sampling.

\subsection{Additional Baseline Approaches}

Apart from the product of experts, we also evaluate other approaches to combine various data modalities. All the evaluated approaches are illustrated in Figure \ref{fig:KG-example-and-queries}.

\subsubsection{Concatenation}

Given pairs of aligned entities, each pair is characterized by a single vector wherein all modality features of both entities are concatenated. For each pair of aligned entities we create a number of negative alignments, each of which is also characterized by a concatenation of all modality features of both entities. A logistic regression is trained taking these vectors as input, and their corresponding class label (+1 and -1 for positive and negative alignments, respectively). The output of the logistic regression indicates the posterior probability of two entities being the same. In Section \ref{sec:experiments} we refer to this approach as \textsc{Concat}.

\subsubsection{Ensemble}

The ensemble approach combines the various expert models into an ensemble classifier. Instead of training the experts jointly and end-to-end, here each of the expert models is first trained independently. At test time, the scores of the expert models are added and used to rank the entities. We refer to this approach as \textsc{Ensemble}.


\section{Experiments}
\label{sec:experiments}

We conducted experiments on two pairs of knowledge graphs of \mmkg, namely, (\textsc{FB15k} \textit{vs.} \textsc{DB15k} and \textsc{Yago15k} \textit{vs.} \textsc{FB15k}). We evaluate a number of different instances of the product of experts (PoE) model, as well as the other baseline methods, in the $\mathtt{sameAs}$ prediction task. Because of its similarity with link prediction, we use metrics commonly used for this task. The main objective of the experiments is to demonstrate that \mmkg is suitable for the task at hand, and specifically that the related problems can benefit from learning of multiple feature types.

\subsection{Evaluation}
\mmkg allows to experiment with different percentages of aligned entities between KGs. These alignments are given by the $\mathtt{sameAs}$ predicates that we previously found. We evaluate the impact of the different modalities in scenarios wherein the number of given alignments $P$ [$\%$] between two KGs is low, medium and high. We reckon that such scenarios would correspond to 20\%, 50\% and 80\% out of all $\mathtt{sameAs}$ predicates, respectively. We use these alignments along with the two KGs as part of our observed triples $\mathbf{T}$, and split equally the remaining $\mathtt{sameAs}$ triples into validation and test.

\begin{table}[]
	\centering
		\caption{\label{numerical} $\mathtt{sameAs}$ queries for which numerical experts led to good performance. Left and right column correspond to \textsc{FB15k} and \textsc{DB15k}, respectively.}
	\scalebox{0.9}{
	\begin{tabular}{|cc|cc|}
		\hline
		\multicolumn{2}{|c|}{/m/015dcj}          & \multicolumn{2}{c|}{Marc\_Christian}           \\ \hline
		date\_of\_birth               & 1925.11 & birthDate                  & 1925.11           \\
		date\_of\_death               & 1985.10 & deathDate                  & 1985.10           \\
		height\_meters                & 1.93    & height                     & 1.9558            \\ \hline
		\multicolumn{2}{|c|}{/m/07zhjj}          & \multicolumn{2}{c|}{How\_i\_met\_your\_mother} \\ \hline
		number\_of\_seasons           & 9.0     & numberOfSeasons            & 9.0               \\
		air\_date\_of\_final\_episode & 2014.03 & completionDate             & 2014.03           \\
		number\_of\_episodes          & 208.0   & numberOfEpisodes           & 208.0             \\ \hline
	\end{tabular}
}
\end{table}

\begin{figure}[!tbp]
  \caption{\label{images} $\mathtt{sameAs}$ queries for which visual experts led to good performance. Left and right images within each pair correspond to \textsc{FB15k} and \textsc{DB15k}, respectively.}
  \begin{minipage}[b]{0.5\textwidth}
    \begin{center}
    \includegraphics[scale=0.25]{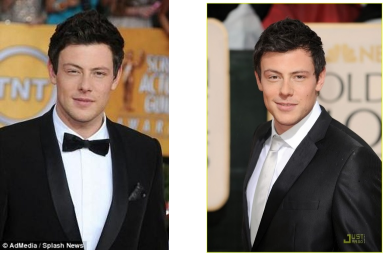}
    \end{center}
  \end{minipage}
  \hfill
  \begin{minipage}[b]{0.5\textwidth}
    \begin{center}
    \includegraphics[scale=0.25]{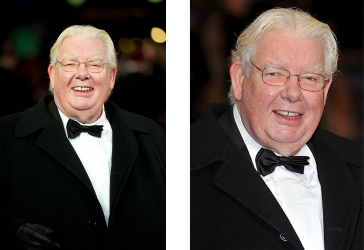}
    \end{center}
  \end{minipage}
\end{figure}

We use \textsc{Amie+}~\cite{Galarraga:2015} to mine relational features for the relational experts. We used the standard settings of \textsc{Amie+} with the exception that the minimum absolute support was set to $2$ and the maximum number of entities involved in the rule to four. The latter is important to guarantee that \textsc{Amie+} retrieves rules like $(x, \mathtt{r_1}, w), (w, \mathtt{SameAs}, z), (z, \mathtt{r_2}, y) \Rightarrow (x,\mathtt{SameAs}, y)$, wherein $\mathtt{r_1}$ is a relationship that belongs to the one KG, and $\mathtt{r_2}$ to the other KG. One example of retrieved rule by \textsc{AMIE+} is:

\begin{center}
$(x, \mathtt{father\_of_{DB15k}}, w), (w, \mathtt{SameAs}, z), (z, \mathtt{children\_of_{FB15k}}, y) \Rightarrow (x,\mathtt{SameAs}, y)$
\end{center}

In this case both $\mathtt{father\_of_{DB15k}}$ and $\mathtt{children\_of_{FB15k}}$ are (almost) functional relationships. A relationship $\mathtt{r}$ is said to be functional if an entity can only be mapped exactly to one single entity via $\mathtt{r}$. The relational expert will learn that the body of this rule leads to a $\mathtt{sameAs}$ relationship between entities $x$ and $y$ and with a very high likelihood.

We used \textsc{Adam} \cite{KingmaB14} for parameter learning in a mini-batch setting with a learning rate of $0.001$, the categorical cross-entropy as loss function and the number of epochs was set to 100. We validated every 5 epochs and stopped learning whenever the MRR (Mean Reciprocal Rank) values on the validation set decreased. The batch size was set to $512$ and the number $N$ of negative samples to 500 for all experiments.

\begin{table*}[!htbp]
\centering
\caption{$\mathtt{sameAs}$ prediction on \textsc{FB15K}-\textsc{DB15K} for different percentages of $P$.}
\label{tab:results_DB_FB}
\scalebox{0.85}{
	\begin{tabular}{|c|c|c|c|c|c|c|c|c|c|}
		\hline
		$P$ [$\%$]        & \multicolumn{3}{c|}{20\%}                     & \multicolumn{3}{c|}{50\%}                     & \multicolumn{3}{c|}{80\%}                     \\ \hline
							  & MRR           & Hits@1        & Hits@10       & MRR           & Hits@1        & Hits@10       & MRR           & Hits@1        & Hits@10       \\ \hline
		PoE-\textit{n}                   & 12.8          & 10.1          & 18.6          & 23.0          & 16.8          & 34.1          & 28.2          & 21.8          & 37.4          \\ \hline
		PoE-\textit{r}                   & 12.9		  & 10.7  		  & 16.5          & 26.3		  & 22.9  		  & 31.7          & 35.9		  & 33.6 		  & 38.6          \\ \hline
		PoE-\textit{l}                & 12.2          & 7.9           & 20.3          & 42.8          & 34.9          & 58.2          & 63.1          & 55.6          & 76.6          \\ \hline
		PoE-\textit{i}                   & 1.6           & 0.8           & 2.7           & 2.3           & 1.3           & 3.8           & 3.3           & 1.7           & 5.9           \\ \hline
		PoE-\textit{lni}        & 16.7          & 12.0          & 25.6          & 48.1          & 40.9          & 62.1          & 68.5          & 62.0          & 79.3          \\ \hline
		PoE-\textit{rni}           & \textbf{28.3} & \textbf{23.2} & \textbf{39.0} & 44.2		  & 38.0 		  & 55.7 		  & 55.8 		  & 50.2 		  & 64.1          \\ \hline
		PoE-\textit{lri}        & 12.5 		  & 8.8  		  & 19.1          & 40.4 		  & 33.4 		  & 53.9          & 67.0 		  & 60.3 		  & 78.3          \\ \hline
		PoE-\textit{lrn}        & 16.0 		  & 11.5 		  & 24.1          & 47.9		  & 41.2 		  & 60.3          & 67.1 		  & 60.6 		  & 79.1          \\ \hline
		PoE-\textit{lrni}    & 17.0 		  & 12.6 		  & 25.1		  & \textbf{53.3} & \textbf{46.4} & \textbf{65.8} & \textbf{72.1} & \textbf{66.6} & \textbf{82.0}          \\ \hline
	\end{tabular}
}
\end{table*}

\begin{table*}[!htbp]
	\centering
	\caption{$\mathtt{sameAs}$ prediction on \textsc{FB15K}-\textsc{Yago15K} for different percentages of $P$.}
	\label{tab:results_Yago_FB}
	\scalebox{0.85}{
	\begin{tabular}{|c|c|c|c|c|c|c|c|c|c|}
		\hline
		$P$ [$\%$]               & \multicolumn{3}{c|}{20\%}                     & \multicolumn{3}{c|}{50\%}                     & \multicolumn{3}{c|}{80\%}                     \\ \hline
							  & MRR           & Hits@1        & Hits@10       & MRR           & Hits@1        & Hits@10       & MRR           & Hits@1        & Hits@10       \\ \hline
		PoE-\textit{n}                   & 22.2          & 15.4          & 33.7          & 38.9          & 29.7          & 56.1          & 35.8          & 27.1          & 53.5          \\ \hline
		PoE-\textit{r}                   & 9.9  		  & 8.4           & 12.3          & 20.0 		  & 18.0          & 23.1          & 29.9 		  & 28.1          & 31.9          \\ \hline
		PoE-\textit{l}                & 10.1          & 6.4           & 16.9          & 32.0          & 25.8          & 44.1          & 50.5          & 43.7          & 63.6          \\ \hline
		PoE-\textit{i} 			      & 1.4           & 2.4           & 0.7           & 2.0           & 1.1           & 3.2           & 3.2           & 1.7           & 5.5           \\ \hline
		PoE-\textit{lni}        & 15.4          & 10.9          & 24.1          & 39.8          & 32.8          & 52.6          & 59.0          & 52.5          & 70.5          \\ \hline
		PoE-\textit{rni}           & \textbf{33.4} & \textbf{25.0} & \textbf{49.5} & \textbf{49.8} & \textbf{41.1} & \textbf{66.9} & 57.2		  & 49.2 		  & 70.5          \\ \hline
		PoE-\textit{lri}        & 11.3 		  & 7.7  		  & 18.1          & 34.0		  & 28.1 		  & 44.7          & 55.5 		  & 49.3 		  & 66.7          \\ \hline
		PoE-\textit{lrn}        & 13.9 		  & 10.2 		  & 20.9          & 37.3		  & 31.6 		  & 47.4          & 57.7		  & 51.3  		  & 68.9          \\ \hline
		PoE-\textit{lrni}    & 15.4 		  & 11.3 		  & 22.9		  & 41.4		  & 34.7 		  & 53.6		  & \textbf{63.5} & \textbf{57.3} & \textbf{74.6}  \\ \hline
	\end{tabular}
}
\end{table*}


We follow the same evaluation procedure as previous works of the link prediction literature. Therefore, we measure the ability to answer completion queries of the form $(\mathtt{h}, \mathtt{SameAs}, \mathtt{t?})$ and $(\mathtt{h?}, \mathtt{SameAs}, \mathtt{t})$.  For queries of the form $(\mathtt{h}, \mathtt{SameAs}, \mathtt{t?})$, wherein $\mathtt{h}$ is an entity of the first KG, we replaced the tail by each of the second KB's entities in turn, sorted the triples based on the scores or probabilities, and computed the rank of the correct entity. We repeated the same process for the queries of type $(\mathtt{h?}, \mathtt{SameAs}, \mathtt{t})$, wherein $\mathtt{t}$ in this case corresponds to an entity of the second KG and we iterate over the entities of the first KG to compute the scores. The mean of all computed ranks is the Mean Rank (lower is better) and the fraction of correct entities ranked in the top $n$ is called hits@$n$ (higher is better). We also computer the Mean Reciprocal Rank (higher is better) which is an evaluation metric that is less susceptible to outliers. Note that the filtered setting described in \cite{bordes2013translating} does not make sense in this problem, since an entity can be linked to an entity via a $\mathtt{SameAs}$ relationship only once.

We report the performance of the PoE in its full scope in Tables \ref{tab:results_DB_FB} and \ref{tab:results_Yago_FB}. We also show feature ablation experiments, each of which corresponds to removing one modality from the full set. The performance of each modality in isolation is also depicted. We use the abbreviations PoE-\textit{suffix} to refer to the different instances of PoE{.} \textit{suffix} is a combination of the letters \textsc{l} (\textsc{l}atent), \textsc{r} (\textsc{r}elational), \textsc{n} (\textsc{n}umerical) and \textsc{i} (\textsc{i}mage) to indicate the inclusion of each of the four feature types. Generalizations are complicated to make, given that performance of PoE's instances differ across percentages of aligned entities and pairs of knowledge graphs. Nevertheless,  there are two instances of our PoE approach, PoE-\textit{lrni} and PoE-\textit{rni}, that tend to outperform all others for low and high percentages of aligned entities, respectively. Results seem to indicate that the embedding expert response dominates over others, and hence its addition to PoE harms the performance when such expert is not the best-performing one. Table \ref{numerical} and Figure \ref{images} provides examples of queries where numerical and visual information led to good performance, respectively. It is hard to find one specific reason that explains when adding numerical and visual information is beneficial for the task at hand. For example, there are entities with a more canonical visual representation than others. This relates to the difficulty of learning from visual data in the $\mathtt{sameAs}$ link prediction problem, as visual similarity largely varies across entities. Similarly, the availability of numerical attributes largely varies even for entities of the same type within a KG. However, Tables  \ref{tab:results_DB_FB} and \ref{tab:results_Yago_FB} provide empirical evidence of the benefit from including additional modalities.

Table \ref{combinationComparison} depicts results for the best-performing instance of PoE and baselines discussed in Section \ref{sec:methods}. The best performing instance of PoE significantly outperforms the approaches \textsc{Concat} and \textsc{Ensemble}. This validates the choice of the PoE approach, which can incorporate data modalities to the link prediction problem in a principled manner.

\begin{table}[t]
	\centering
	\caption{Performance comparison for  $P=80\%$.}
	\label{combinationComparison}
	\scalebox{0.9}{
	\begin{tabular}{@{}ccccc@{}}
		\toprule
		&                & MRR      & HITS@1     & HITS@10     \\ \midrule
		\multirow{3}{*}{\textsc{FB}-\textsc{DB}}   & \textsc{Concat}  &  2.1     & 1.7     &  2.7          \\
		& \textsc{Ensemble}  	    					   &  40.1    & 34.3    &  50.2         \\
		& PoE-\textit{lrni} 					   &  \textbf{72.1}    & \textbf{66.6}    &  \textbf{82.0}         \\ \midrule
		\multirow{3}{*}{\textsc{FB}-\textsc{Yago}} & \textsc{Concat}  &  0.18    & 0.18    &  0.04         \\
		& \textsc{Ensemble}     						   &  47.6    & 42.3    &  57.5      	\\
		& PoE-\textit{lrni} 					   &  \textbf{63.5}    & \textbf{57.3}    &  \textbf{74.6}         \\ \bottomrule
	\end{tabular}
}
\end{table}

\section{Conclusion}
\label{sec:conclusion}
We present \mmkg, a collection of three knowledge graphs that contain multi-modal data, to benchmark link prediction and entity matching approaches. An interesting property of \mmkg is that the three knowledge graphs are very heterogeneous with respect to the number of relation types and the degree of sparsity, for instance. An extensive set of experiments validate the utility of the data set in the $\mathtt{sameAs}$ link prediction task.

\bibliographystyle{splncs04}
\bibliography{ref}

\end{document}